\documentclass{article}

\usepackage{PRIMEarxiv}

\usepackage[utf8]{inputenc} 
\usepackage[T1]{fontenc}    
\usepackage{hyperref}       
\usepackage{url}            
\usepackage{booktabs}       
\usepackage{amsfonts}       
\usepackage{nicefrac}       
\usepackage{microtype}      
\usepackage{lipsum}
\usepackage{fancyhdr}       
\usepackage{graphicx}       
\graphicspath{{media/}} 
\usepackage{multirow}

\title{WavePaint: Resource-efficient Token-mixer for Self-supervised Inpainting
}

\author{
  Pranav Jeevan, Dharshan Sampath Kumar, Amit Sethi \\
  Department of Electrical Engineering \\
  Indian Institute of Technology Bombay \\
  Mumbai, India\\
  \texttt{\{pranav13phoenix, dharshan2609 \}@gmail.com} \\
}

\begin{document}
\maketitle

\begin{abstract}
Image inpainting, which refers to the synthesis of missing regions in an image, can help restore occluded or degraded areas and also serve as a precursor task for self-supervision. The current state-of-the-art models for image inpainting are computationally heavy as they are based on transformer or CNN backbones that are trained in adversarial or diffusion settings. This paper diverges from vision transformers by using a computationally-efficient WaveMix-based fully convolutional architecture -- WavePaint. It uses a 2D-discrete wavelet transform (DWT) for spatial and multi-resolution token-mixing along with convolutional layers. The proposed model outperforms the current state-of-the-art models for image inpainting on reconstruction quality while also using less than half the parameter count and considerably lower training and evaluation times. Our model even outperforms current GAN-based architectures in CelebA-HQ dataset without using an adversarially trainable discriminator. Our work suggests that neural architectures that are modeled after natural image priors require fewer parameters and computations to achieve generalization comparable to transformers.
\end{abstract}


\begin{figure}[b]
\centering
\includegraphics[scale=0.5]{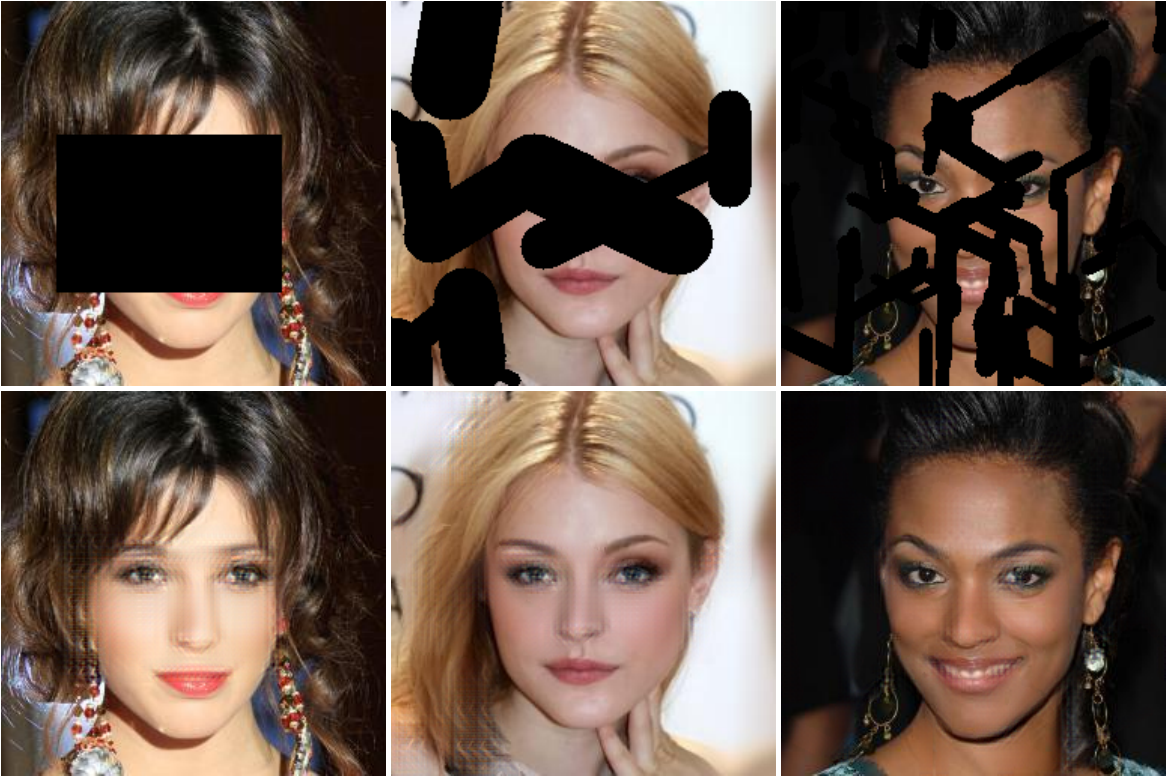}
\caption{A sample of inpainted images (bottom row) generated by WavePaint from masked images (top row) from CelebA-HQ set using wide, medium, and narrow masks, respectively}
\label{fig:pics}
\end{figure}

\section{Introduction}
\label{sec:intro}

Image inpainting refers to the process of filling of missing parts of an image (blemishes, holes, and other defects) realistically to match the available context, thereby restoring a degraded image. It requires implicitly modeling large scale structures in natural images and an ability to perform image synthesis. State-of-the-art inpainting models are based on deep neural networks trained in a self-supervised and adversarial manner by automatically generating training samples from large image datasets by randomly masking parts of the image.

\begin{figure}[]
\centering
\includegraphics[scale=0.4]{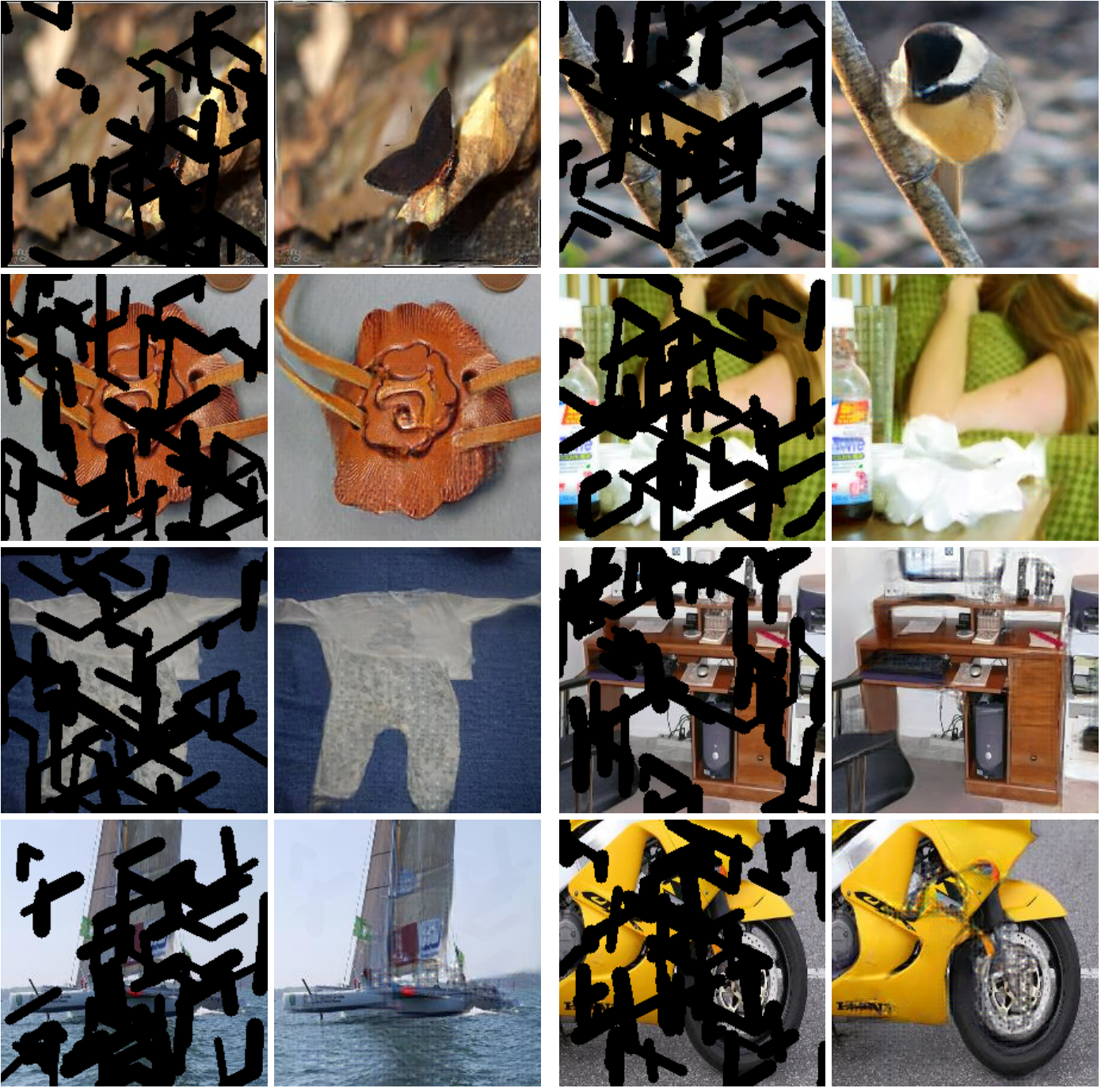}
\caption{Inpainted images (second and fourth columns) generated by WavePaint from masked images (first and third columns) from ImageNet validation set}
\label{fig:pics}
\end{figure}

Some image reconstruction tasks, such as inpainting with large masks, require networks to have large effective receptive fields~\cite{luo2017understanding}. Convolutional neural networks (CNN) require deep architectures (a large number of layers) for increasing the receptive fields. On the other hand, using self-attention to access all the pixels of an image right from the first layer gives transformers large receptive fields. However, the quadratic complexity with respect to sequence length (number of patches) introduces an enormous computational burden on transformers. Moreover, transformers require larger training data than CNNs since they lacks the inductive bias of spatial equivariance.

The search for efficient models that can mix global spatial information while retaining the inductive bias of CNNs has led to the development of token-mixing models such as PoolFormer~\cite{yu2022metaformer}, ConvMixer~\cite{trockman2022patches} and WaveMix~\cite{jeevan2023wavemix} which use pooling, depth-wise convolutions and 2D-discrete wavelet transform (2D-DWT), respectively.  These alternatives consume a fraction of the resources compared to transformers to achieve competitive generalization in tasks such as classification and segmentation. The performance of these models on image generation or restoration tasks has not been evaluated. 

Our model is a neural architecture that is inspired by WaveMix~\cite{jeevan2023wavemix} and ConvMixer~\cite{trockman2022patches}. 
We investigated the application of WaveMix architectural framework to the task of image inpainting with suitable adaptations to the previously proposed architectures.
This choice is motivated by the success of WaveMix in approaching the state-of-the--art (SOTA) for different datasets on the task of parameter-efficient image classification and segmentation by modeling additional inductive priors of images, such as scale invariance.

Specifically, we have worked on large mask inpainting, where the mask occlude a substantial and non-trivial part of the image, but its shape is known. We have not worked on blind mask inpainting where the model does not see the mask. Sending mask to the model is necessary in the large-mask setting for the model to know where the mask is and where to fill information.

Our contributions are summarized below:

\begin{itemize}
    \item We present -- WavePaint -- a token-mixing network modeled after natural image priors that can perform image inpainting. The network is based on recently proposed WaveMix architecture which uses 2D-discrete wavelet transform for spatial token-mixing. We also employ depth-wise convolution in our network for additional token-mixing. The presence of spatial token-mixing enables the model to have faster receptive field expansion compared to CNNs, which helps in better image reconstruction through access to global context of the image.
    
    \item The use of a paramter-free 2D-DWT and parameter-efficient depth-wise convolution helps WavePaint reconstruct images without the need for large number of model parameters. WavePaint with 5M parameters can outperform much larger models such as LaMa(27 M) and CoModGAN (109 M)~\cite{CoMod_GAN2021} on CelebA-HQ dataset in multiple mask sizes. It is able to achieve these results consuming less resources and time.
    
    \item WavePaint does not need advesarial or diffusion based training routines, which are slow. The ability of wavelet token mixing to generate realistic images from masked ones shows that we can develop more efficient neural networks for image generation. 
    
    \item Complicated multi-stage models have been proposed that generate intermediate predictions which are further processed to restore the missing parts~\cite{liu2020rethinking, nazeri2019edgeconnect, song2018spgnet}. Our model reconstructs the image using a simple single-stage network.

    \item We show that utilizing natural image priors in neural architectural design may be the way forward to avoid large computational costs and training datasets.

\end{itemize}

\section{Related Works}

Mask-Aware Dynamic Filtering (MADF)~\cite{Zhu_2021} uses an encoder-decoder framework  to learn multi-scale features for missing regions in the encoding phase. It adopts Point-wise Normalization (PN) in decoding phase by considering the statistical nature of features at masked points. It does not use adversarial training using a discriminator.

\subsection{Generative Adversarial Networks}

Co-ModGAN ~\cite{CoMod_GAN2021} is a GAN model  which introduces variability into the generated outputs by integrating input image-conditional and unconditional generators. It combines an unconditional style vector with an input-conditioned style vector through a linear transformation into a single modulated output. The conditional vector is obtained from an encoder network, and the unconditional vector is obtained by passing a noise vector through a pre-trained FCN, as done in Style GAN ~\cite{stylegan2018}. Finally, this combined output is passed through the decoder to generate the output. 

Image completion with transformer (ICT)~\cite{wan2021highfidelity} is a transformer -CNN hybrid model that uses transformers to model the long-range relationships in images to recover pluralistic coherent structures together with coarse textures, and uses CNN for texture replenishment.

Mask-Aware Transformer~\cite{li2022mat} uses a multi-head contextual attention for long-range dependency modeling by exploiting valid tokens indicated by a dynamic mask for directly processing high-resolution images. It also proposed a modified transformer block to increase the stability of large mask training. 

LaMa~\cite{Lama_2021} uses Fast Fourier Convolution (FFC) blocks to understand the local and global context of an image. The use of FFC helps in having an image-wide receptive field. The use of FFC can be considered as a token-mixing operation similar to WaveMix where fast fourier transform is used for spatial token-mixing. It also uses a high receptive field perceptual loss and large training masks. 
 
WaveFill~\cite{yu2021wavefill} uses 2D-DWT to decompose images into multiple frequency bands and fills the missing regions in each frequency band separately. It applies $L1$ reconstruction loss to the decomposed low-frequency bands and adversarial loss
to high-frequency bands to mitigate inter-frequency conflicts and also uses a normalization scheme to align multi-frequency features.

\subsection{Diffusion Models}

Diffusion models uses a T fold pass through a fixed network to go from completely random noise to a coherent and contextually consistent image. Even though the overall performance of diffusion models are excellent, the training and inference process are extremely time-consuming. 

Latent diffusion model (LDM)~\cite{rombach2021highresolution} works on a lower-dimensional feature space rather than the image space to address the time-consuming nature of diffusion training. The model uses an encoder-decoder architecture with the slow diffusion step at the neck of the chain to speed up the entire network. 

RePaint~\cite{lugmayr2022repaint} is a denoising diffusion probabilistic model (DDPM) based inpainting approach which employs a pretrained unconditional DDPM as the generative prior. It only alters the reverse diffusion iterations by sampling the unmasked area to condition the generation process. Additionally, they perform re-sampling on the generated output at each step, by noising and successively de-noising a fixed number of times, in order to get a coherent image. Thus. the model produces high quality and diverse output images for any masked images.

\section{WavePaint Architectural Framework}
\label{sec:3}

\begin{figure*}[t]
\centering
\includegraphics[scale=0.7]{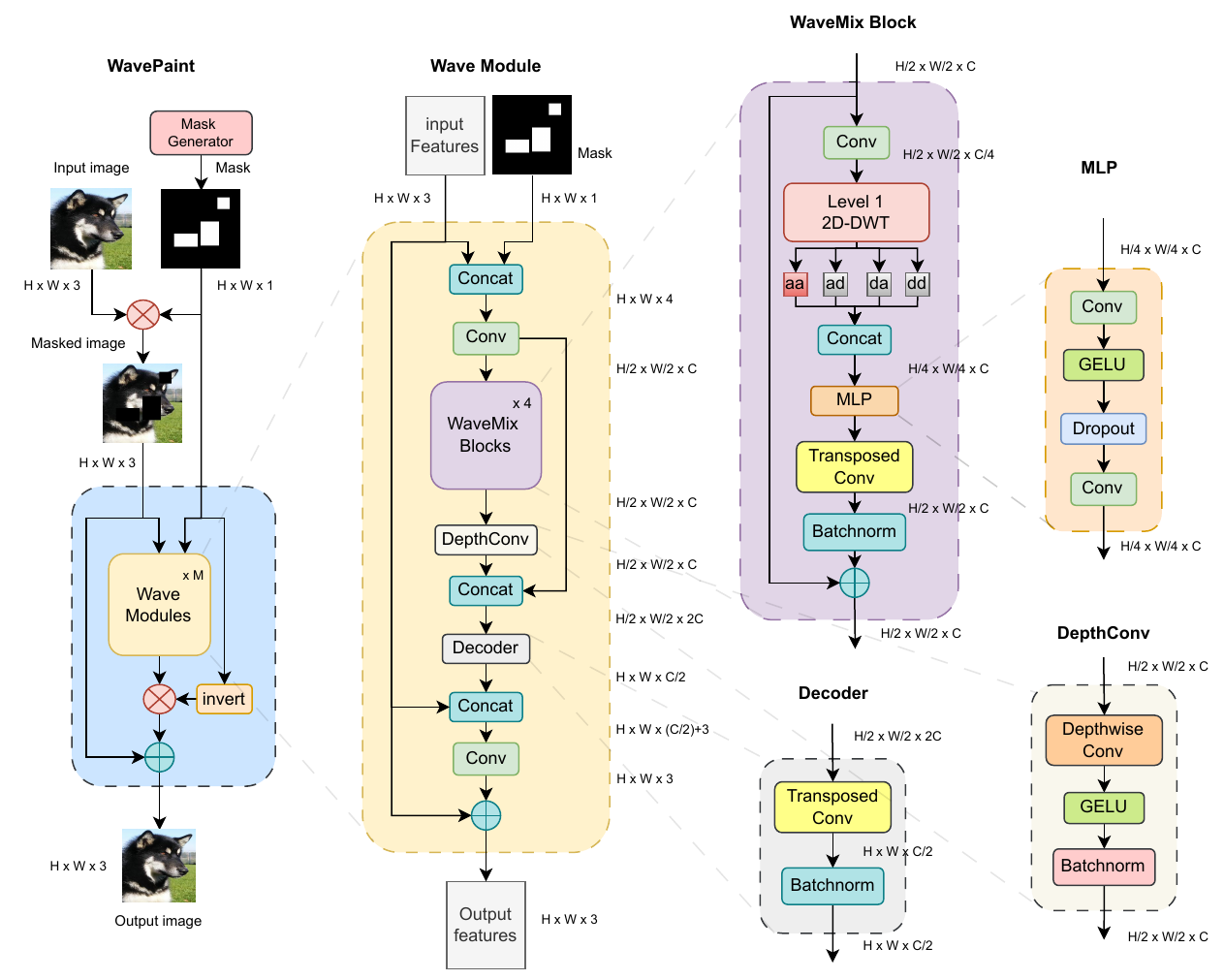}
\caption{Architecture of WavePaint along with details of Wave module, WaveMix block, Decoder, DepthConv and MLP are shown. The resolutions of feature maps after each operation is also provided for an input of $H\times W\times 3$. WaveMix block is taken from~\cite{jeevan2023wavemix}}
\label{fig:wavepaint}
\end{figure*}

Observing the success of WaveMix and ConvMixer which uses 2D-DWT and depthwise-convolutions respectively for parameter efficient token-mixing, we have created a neural architecture that can inpaint masked images using these token-mixing operations. The ability of these token-mixers to impart rapid receptive field expansion from initial layers itself helps the model grasp the global context faster than conventional CNN-based networks. 
Unlike other popular models for image inpainting that uses diffusion or adversarial training, our model has simple single network architecture and can perform well without the need for a discriminator network.

\subsection{Overall architecture}

The input image $\textbf{x}\in\mathbb{R}^{H\times W \times 3}$ is masked by a binary mask $m\in\mathbb{R}^{H\times W \times 1}$ that is generated from a mask generator. The masked image is denoted as $\textbf{x} \oplus m$. The mask $m$ is concatenated with the masked image $\textbf{x} \oplus m$, resulting in a 4-channel input $\hat{\textbf{x}}\in\mathbb{R}^{H\times W \times 4}$ that is passed to the model as shown in Figure~\ref{fig:wavepaint}.

The network consists of a series of $M$ Wave modules which processes the input $\hat{\textbf{x}}$ and gives the output $\hat{\textbf{y}}\in\mathbb{R}^{H\times W \times 3}$. $\hat{\textbf{y}}$ is multiplied by the inverted binary mask, $1 - m$ to hide the unmasked areas of the output and retains the inpainted parts by the model. This is added back to the masked image $\hat{\textbf{x}}$ which fills the unmasked areas and creates the final inpainted image $\textbf{y}\in\mathbb{R}^{H\times W \times 3}$. This ensures that the model only fills the masks areas and not change pixel information of unmasked parts.

\subsection{Wave Modules}

Proper inpainting requires global context information of the image. WaveMix has shown rapid expansion of receptive fields from very early layers~\cite{jeevan2023wavemix}. So we use $4$ WaveMix blocks in series in each of the Wave modules to process the image and get global context. This is further aided by the depth-wise convolution layer which further helps with spatial token-mixing with high parameter-efficiency~\cite{trockman2022patches}.

Denoting input and output tensors of the Wave module by $\hat{\textbf{x}}_{in}$ and $\hat{\textbf{x}}_{out}$, respectively; convolution operations by $c_1$ and $c_2$ and its respective trainable parameter sets by $\theta_1$ and $\theta_2$ respectively;  the series of WaveMix blocks by $WB$; DepthConv by $DC$; Decoder by $D$;  concatenation along the channel dimension by $\oplus$, and point-wise addition by $+$, the operations inside a Wave module can be expressed using the following equations:

\begin{equation} \label{eq:1}
    \hat{\textbf{x}}_0 = \hat{\textbf{x}}_{in} \oplus m ; \hspace{2cm}    
    \hat{\textbf{x}}_{in}\in\mathbb{R}^{H\times W \times 4}     
\end{equation}

\begin{equation} \label{eq:2}
    \hat{\textbf{x}}_{1} = c_1(\hat{\textbf{x}}_{0}, \theta_1) ; \hspace{1.8cm}    
    \hat{\textbf{x}}_{1}\in\mathbb{R}^{H/2\times W/2 \times C}     
\end{equation}

\begin{equation}\label{eq:3}
    \hat{\textbf{x}}_{2} = WB(\hat{\textbf{x}}_{1}) ; \hspace{2cm}   
    \hat{\textbf{x}}_{2}\in\mathbb{R}^{H/2\times W/2 \times C}     
\end{equation}

\begin{equation} \label{eq:4}
    \hat{\textbf{x}}_{3} = DC(\hat{\textbf{x}}_{2}) ; \hspace{2cm}   
    \hat{\textbf{x}}_{3}\in\mathbb{R}^{H/2\times W/2 \times C}     
\end{equation}

\begin{equation} \label{eq:5}
    \hat{\textbf{x}}_{4} = \hat{\textbf{x}}_{3} \oplus \hat{\textbf{x}}_{1} ; \hspace{2cm}   
    \hat{\textbf{x}}_{4}\in\mathbb{R}^{H/2\times W/2 \times 2C}     
\end{equation}

\begin{equation} \label{eq:6}
    \hat{\textbf{x}}_{5} = D(\hat{\textbf{x}}_{4}) ; \hspace{2cm}   
    \hat{\textbf{x}}_{5}\in\mathbb{R}^{H\times W \times C/2}     
\end{equation}

\begin{equation} \label{eq:7}
    \hat{\textbf{x}}_{6} = \hat{\textbf{x}}_{5} \oplus \hat{\textbf{x}}_{in} ; \hspace{1cm}   
    \hat{\textbf{x}}_{6}\in\mathbb{R}^{H\times W \times (C/2+3)}     
\end{equation}

\begin{equation} \label{eq:8}
    \hat{\textbf{x}}_{7} = c_2(\hat{\textbf{x}}_{6}, \theta_2) ; \hspace{1.8cm}    
    \hat{\textbf{x}}_{7}\in\mathbb{R}^{H\times W \times 3}     
\end{equation}

\begin{equation} \label{eq:9}
    \hat{\textbf{x}}_{out} = \hat{\textbf{x}}_{7} + \hat{\textbf{x}}_{in}; \hspace{1.8cm}    
    \hat{\textbf{x}}_{out}\in\mathbb{R}^{H\times W \times 3}     
\end{equation}

Each Wave module receives the input $\hat{\textbf{x}}_{in}\in\mathbb{R}^{H\times W \times 3}$ and the mask $m$ which are concatenated to create $\hat{\textbf{x}}_0$ (\ref{eq:1}). $\hat{\textbf{x}}_0$ is send to a convolution layer $c_{1}$ that reduces its feature resolution  by half and increases the channel dimension to $C$ ( ~\ref{eq:2}). This feature map $\hat{\textbf{x}}_1$ is sent to a series of 4 WaveMix blocks for token-mixing ( ~\ref{eq:3}). The output from the WaveMix block $\hat{\textbf{x}}_2$ is further passed through a DepthConv module where the feature maps undergo further spatial token-mixing from the depth-wise convolution ( ~\ref{eq:4}). A skip connection from $c_{1}$ is concatenated with the output from DepthConv module $\hat{\textbf{x}}_3$ which increases the channel dimension of the output $\hat{\textbf{x}}_4$ to $2C$ ( ~\ref{eq:5}). This output is further passed through a Decoder network which increases the resolution of feature maps to original resolution ( ~\ref{eq:6}). The Decoder layer also reduces the number of channels to $C/2$ and the feature maps $\hat{\textbf{x}}_5$ are again concatenated with the input $\hat{\textbf{x}}_{in}$ ( ~\ref{eq:7}). The output after concatenation $\hat{\textbf{x}}_6$ is then passed to a final convolution layer $c_2$ to generate the output $\hat{\textbf{x}}_7$ ( ~\ref{eq:8}). A residual connection is also provided from the input for ease of gradient flow ( ~\ref{eq:9}) and resultant feature maps are the final output of the Wave module $\hat{\textbf{x}}_{out}$.

\begin{table*}[t]
\centering
\small\addtolength{\linewidth}{-15pt}
\begin{tabular}{@{}lrrrrrrr@{}}
\toprule
\multicolumn{8}{c}{CelebA-HQ ($256\times256$)} \\ \midrule
\multirow{2}{*}{Model} & \multirow{2}{*}{\#Params\textdownarrow} & \multicolumn{2}{l}{Narrow masks} & \multicolumn{2}{l}{Medium masks} & \multicolumn{2}{l}{Wide Masks} \\ \cmidrule(l){3-8} 
 &  & FID\textdownarrow & LPIPS\textdownarrow & FID\textdownarrow & LPIPS\textdownarrow & FID\textdownarrow & LPIPS\textdownarrow \\ \midrule
CoModGAN~\cite{CoMod_GAN2021} & 109 M & 16.8 & \textbf{0.079} & 19.4 & 0.092 & 24.4 & 0.102 \\
AOT GAN~\cite{zeng2021aggregated} & 15 M & 6.67 & 0.081 & 7.28 & 0.089 & 10.3 & 0.118 \\
RegionWise~\cite{ma2019regionwise} & 47 M & 11.1 & 0.124 & 7.52 & 0.101 & 8.54 & 0.121 \\
DeepFill v2~\cite{yu2019freeform} & 4 M & 12.5 & 0.130 & 9.05 & 0.105 & 11.2 & 0.126 \\
EdgeConnect~\cite{nazeri2019edgeconnect} & 22 M & 9.61 & 0.099 & 7.56 & 0.095 & 9.02 & 0.120 \\
LaMa-Fourier~\cite{Lama_2021} & 27 M & 7.26  & 0.085 & 6.13 & \textbf{0.080} & \textbf{6.96} & \textbf{0.098} \\ \midrule
WavePaint &3 M& 8.03 & 0.115 & 8.87 & 0.123 & 21.3 & 0.155\\
WavePaint & 10 M & \textbf{5.53} & \textbf{0.085} & \textbf{5.59} & \textbf{0.090} & \textbf{7.22} & \textbf{0.112} \\\bottomrule
\end{tabular}
\caption{Quantitative evaluation metrics of inpainting on CelebA-HQ dataset. Learned perceptual image patch similarity (LPIPS) and Fr\'echet inception distance (FID) are reported. The best WavePaint results are highlighted in bold. The results of models which report better results than WavePaint are coloured red. The metrics are reported for three different types of test mask generation, i.e. narrow, medium and wide masks as used in LaMa~\cite{Lama_2021}. Other models have much larger parameters compared to WavePaint and also employ adversarial training. Still, WavePaint manages to outperform them without using any adversarial training. It uses learnable parameters more efficiently. }
\label{tab:celeb}
\end{table*}

\subsection{WaveMix Blocks}

WaveMix block~\cite{jeevan2023wavemix} is the fundamental building block of WaveMix architecture which allows multi-resolution token-mixing of information using 2D-DWT. This helps in a rapid expansion of receptive field. It also reduces computational burden because 2D-DWT decreases the input resolution by half and further processing by multi-layer-perceptron (MLP) is faster and cheaper. DWT helps in lowering the number of model parameters significantly, as it lacks any parameters, while promoting global context understanding even on a shallow network. We have used the WaveMix block with one level of 2D-DWT using Haar wavelet. Details of the operations inside WaveMix block are given in \cite{jeevan2023wavemix}. 

\subsection{DepthConv}

DepthConv employs a depth-wise convolution operation followed by a GELU non-linearity and batch-normalization as shown in Figure~\ref{fig:wavepaint}. We use a depth-convolution with kernel size of 5, which is smaller than kernel size used in Convmixer models. This was done further decrease the parameter count. 

\subsection{Decoder}

Decoder module is used to up-sample the resolution of feature maps back to original input resolution to the Wave module. It comprises of a transposed convolution layer followed by a batch-normalization. The transposed convolution layer is also used to reduce the number of channels by 4, from $2C$ to $c/2$.

\section{Experiments and Results}

\subsection{Datasets, Loss Function and Metrics}

We use CelebA-HQ\cite{karras2018progressive} and ImageNet~\cite{5206848} datasets (under MIT Licenses) for our experiments. We use images of size $256\times256$ for CelebA-HQ and $224\times224$ for ImageNet experiments. Validation for each dataset was performed on the entire validation sets of respective datasets. 

We followed the same mask generation policy employed in LaMa~\cite{Lama_2021} and used their settings to generate narrow, medium and wide masks. We took the same 26,000 train images and 2,000 test images from CelebA-HQ that LaMa used for CelebA-HQ experiments. Learned perceptual image patch similarity (LPIPS)~\cite{zhang2018unreasonable} and Fr\'echet inception distance (FID)~\cite{heusel2018gans} are reported as metrics since L1 and L2 distances are not enough to compare inpainted images with large masks where multiple natural completions are possible. Inference throughput on a single GPU was reported in frames/sec (FPS). 

We used a hybrid loss $L_{hybrid}$ to optimize the model parameters. Since we did not employ a discriminator for adversarial training, no adversarial loss was used. We used a weighted sum of $L_1$ (mean absolute error), $L_2$ (mean square error) and $L_{LPIPS}$ as shown below:

\begin{equation}\label{eq:10}
    {L}_{hybrid} = (1 - \alpha){L}_{1} + \alpha{L}_{2} + L_{LPIPS}   
\end{equation}

\subsection{Implementation details}

Due to limited computational resources, the \emph{maximum} number of training epochs was set to 300 for CelebA-HQ and 50 for ImageNet Experiments. All experiments were run on a single 80 GB Nvidia A100 GPU. We used AdamW optimizer ($\alpha = 0.001, \beta_{1} = 0.9, \beta_{2}=0.999, \epsilon = 10^{-8}$) with a weight decay of 0.01 during initial epochs and then used SGD (stochastic gradient descent) with learning rate of $0.001$ and momentum $= 0.9$ during the final 50 epochs~\cite{DBLP:journals/corr/abs-1712-07628, jeevan2022convolutional}. We used the maximum batch-size that could be accommodated in a single GPU for our experiments. We used an embedding dimension ($C$) of 128 in all the Wave modules. Each Wave module has 4 WaveMix blocks unless otherwise specified. 

\begin{table}[]
\centering

\begin{tabular}{@{}lrrcrr@{}}
\toprule
\multirow{2}{*}{Model} & \multirow{2}{*}{\#Param} & \multirow{2}{*}{FID\textdownarrow} & \multirow{2}{*}{GPU} &  \multicolumn{2}{r}{\begin{tabular}[c]{@{}r@{}}Throughput\\ (FPS)\end{tabular}}\\ \cmidrule(l){5-6} 
 &  &  &  & Inference & Train \\ \midrule
LaMa~\cite{Lama_2021} & 27 M & 7.26 & 23 GB & 32 & 11 \\
WavePaint & \textbf{5 M} & \textbf{7.09} & \textbf{11 GB} & \textbf{105} & \textbf{32} \\ \bottomrule
\end{tabular}
\vspace{4mm}
\caption{Comparison of LaMa~\cite{Lama_2021} and WavePaint on parameters, resource-consumption and speed. Results are reported for experiments on CelebA-HQ dataset with narrow masks on single 24 GB RTX 3090 GPU for a batch size of 10. We see that WavePaint is three times faster than LaMa, but consumes only half the GPU resource and uses less than one fifth of LaMa's parameters.}
\label{tab:efficient}
\end{table}
 
\subsection{Results and Discussion}

\begin{table*}[]
\centering
\resizebox{\columnwidth}{!}{%
\begin{tabular}{@{}lrrcrrrrrrrrr@{}}
\toprule
\multirow{2}{*}{Model} & \multirow{2}{*}{\#Modules } & \multirow{2}{*}{\# Blocks} & \multirow{2}{*}{DepthConv } & \multirow{2}{*}{\#Params} & \multicolumn{2}{r}{Narrow masks} & \multicolumn{2}{r}{Medium masks} & \multicolumn{2}{r}{Wide Masks} & \multicolumn{2}{r}{\begin{tabular}[c]{@{}r@{}}Throughput (FPS)\end{tabular}} \\ \cmidrule(l){6-13} 
 &  &  & \multicolumn{1}{l}{} &  & \multicolumn{1}{r}{FID\textdownarrow} & \multicolumn{1}{r}{LPIPS\textdownarrow} & \multicolumn{1}{r}{FID\textdownarrow} & \multicolumn{1}{r}{LPIPS\textdownarrow} & \multicolumn{1}{r}{FID\textdownarrow} & \multicolumn{1}{r}{LPIPS\textdownarrow} & Inference & Train \\ \midrule
\multicolumn{13}{c}{CelebA-HQ (256x256)} \\ \midrule
WavePaint & 2 & 4 & No & 3.3 M & 11.1 & 0.148  & 13.9 & 0.148 & 33.7 & 0.176 & 356 & 165\\
WavePaint & 2 & 4 & Yes & 3.3 M & 8.03 & 0.115 & 8.87 & 0.123 & 21.3 & 0.155 & 322 & 145\\
WavePaint & 3 & 4 & Yes & 5.0 M & 7.09 & 0.103  & 6.96 & 0.104 & 10.2 & 0.131 &  275 & 99\\
WavePaint & 5 & 4 & Yes & 8.4 M & 6.56 & 0.096 & 6.62  & 0.098 & 8.83 & 0.122 &  167& 60 \\
WavePaint & 6 & 4 & Yes & 10 M & \textbf{5.53} & \textbf{0.085} & \textbf{5.59} & \textbf{0.090} & \textbf{7.22} & \textbf{0.112} & 133 & 50 \\ 
             \midrule
\multicolumn{12}{c}{ImageNet ($224\times224$)} \\ \midrule
WavePaint & 2 & 4 & Yes & 3.3 M & 3.26 & 0.134 & 3.72 & 0.108 &-  & - &  333 & 213\\
WavePaint & 3 & 4 & Yes & 5.0 M & 3.21 & 0.138 & 3.47 & 0.106 & - & - & 305  & 126\\ \bottomrule
\end{tabular}
}
\caption{Quantitative evaluation metrics of inpainting by WavePaint of different sizes by varying the number of modules and WaveMix blocks per modules. All models use level-1 2D DWT. The models were evaluated on 2000 images of CelebA-HQ that LaMa used for testing. For evaluation on ImageNet, we used the entire 50,000 images from ImageNet validation set. Inference and training throughput in frames per second (FPS) is reported on a single 80 GB A100 GPU.}
\label{tab:ablation}
\end{table*}

\begin{table*}[]
\centering
\begin{tabular}{@{}lrrcrrrrrrrrr@{}}
\toprule
\multirow{2}{*}{Model} & \multirow{2}{*}{Params} & \multicolumn{2}{r}{Narrow masks} & \multicolumn{2}{r}{Medium masks} & \multicolumn{2}{r}{Wide Masks} & \multicolumn{2}{r}{\begin{tabular}[c]{@{}r@{}}Throughput (FPS)\end{tabular}} \\ \cmidrule(l){3-10} 
  \multicolumn{1}{l}{} &  & \multicolumn{1}{r}{FID\textdownarrow} & \multicolumn{1}{r}{LPIPS\textdownarrow} & \multicolumn{1}{r}{FID\textdownarrow} & \multicolumn{1}{r}{LPIPS\textdownarrow} & \multicolumn{1}{r}{FID\textdownarrow} & \multicolumn{1}{r}{LPIPS\textdownarrow} & Inference & Train \\ \midrule
Level 1 & 5.0 M& \textbf{7.09} & 0.103  & \textbf{6.96} & 0.104 & 10.2 & 0.131 &  275 & 99\\
Level 2 & 7.6 M& 7.12 & 0.095 & 7.16 &  0.096 &  \textbf{9.16} & 0.119 & 222  & 78\\
Level 3 & 10 M&  7.74&  \textbf{0.094} & 7.62 & \textbf{0.092} & 9.26 & \textbf{0.112} & 200 &67\\ \bottomrule
\end{tabular}
\caption{The variation of performance of WavePaint with different levels of 2D -DWT used in WaveMix blocks. WavePaint with 3 modules and 4 WaveMix blocks per module is used and results on CelebA-HQ are shown. The improved performance is due to the rapid expansion of receptive fields while using multi-level 2D-DWT token-mixing.}
\label{tab:levels}
\end{table*}

\subsection{Quantitative Results}
\begin{figure*}[t]
\centering
\includegraphics[scale=0.37]{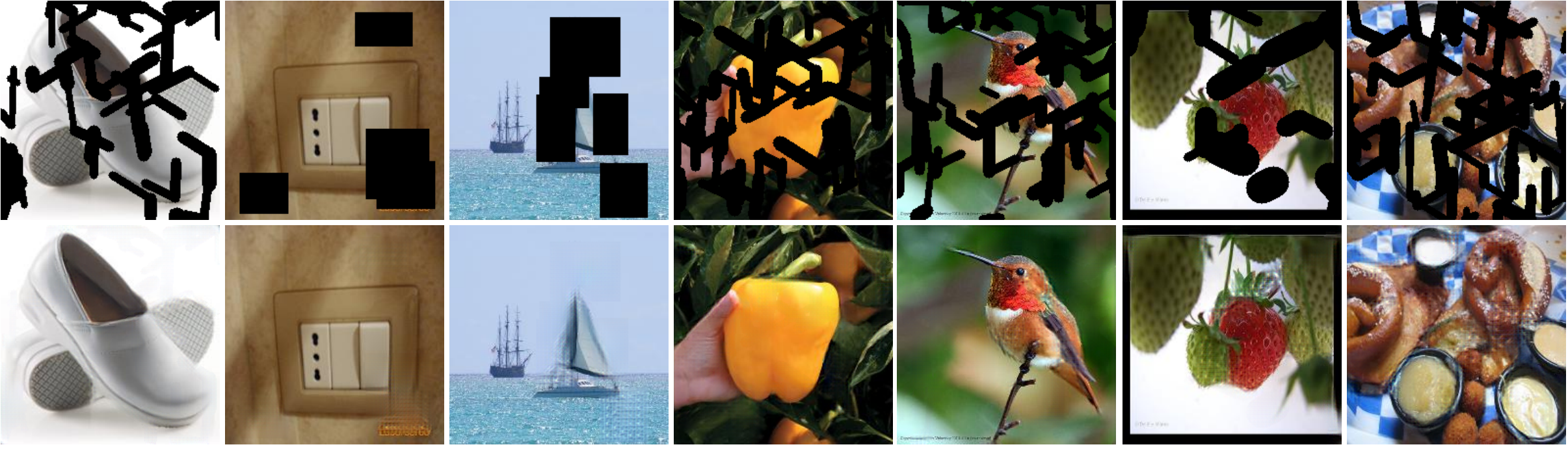}
\caption{Inpainted images (bottom row) generated by WavePaint from masked images (top row) from ImageNet validation set}
\label{fig:picsimage}
\end{figure*}

\begin{figure*}[t]
\centering
\includegraphics[scale=0.42]{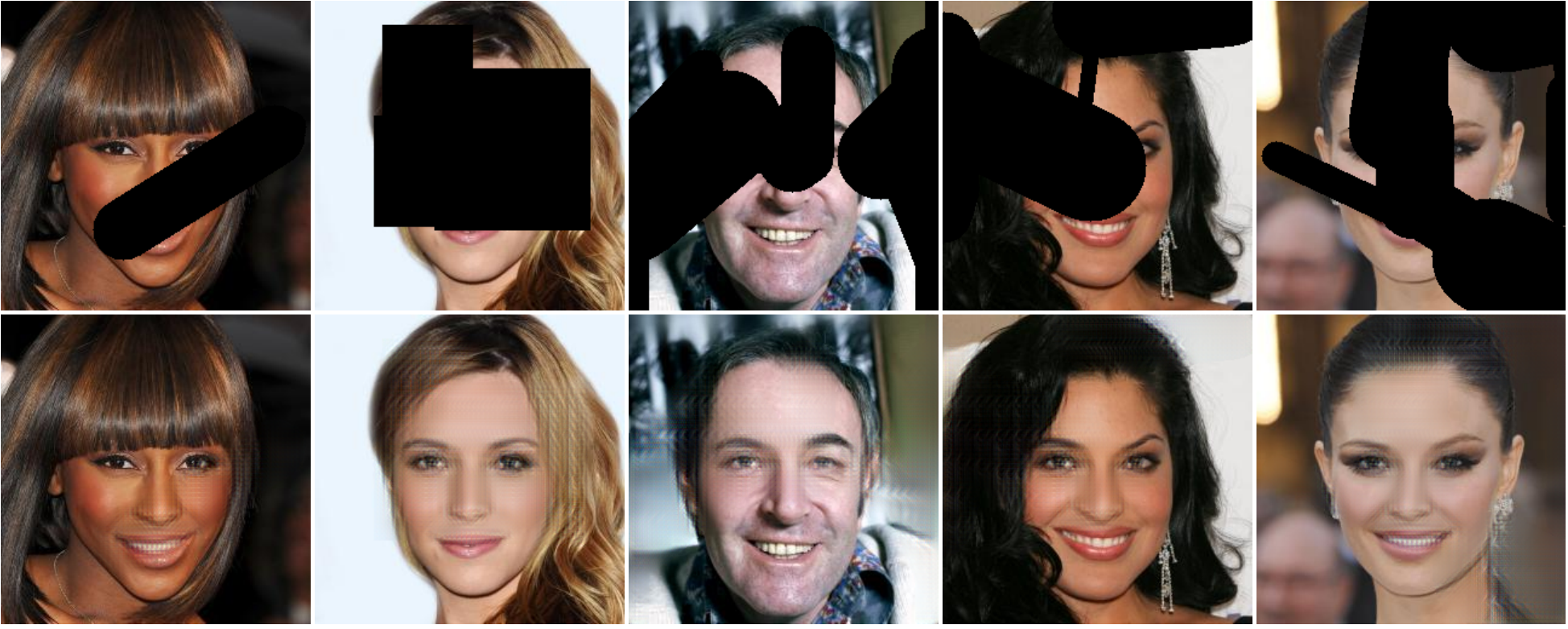}
\caption{Inpainted images (bottom row) generated by WavePaint from masked images using wide masks (top row) from ImageNet validation set}
\label{fig:wide}
\end{figure*}

\begin{figure*}[t]
\centering
\includegraphics[scale=0.42]{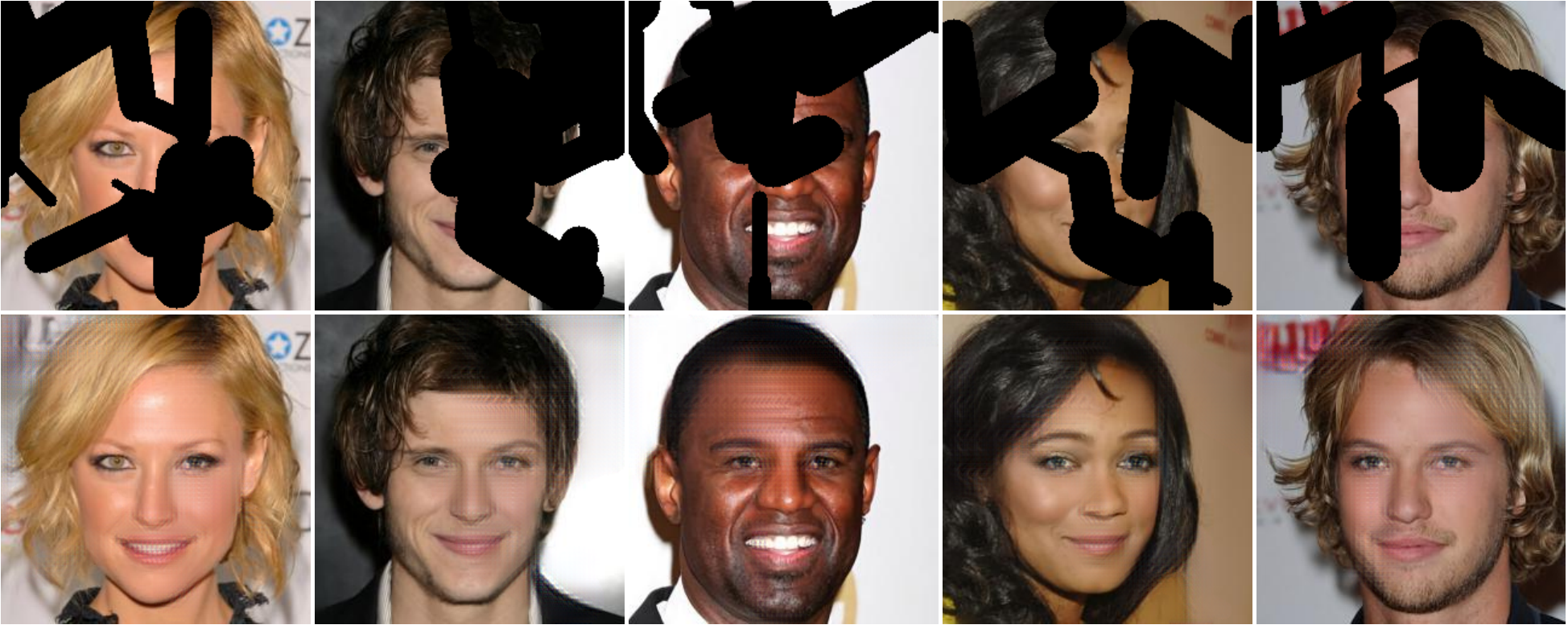}
\caption{Inpainted images (bottom row) generated by WavePaint from masked images using medium masks (top row) from ImageNet validation set}
\label{fig:medium}
\end{figure*}

\begin{figure*}[]
\centering
\includegraphics[scale=0.42]{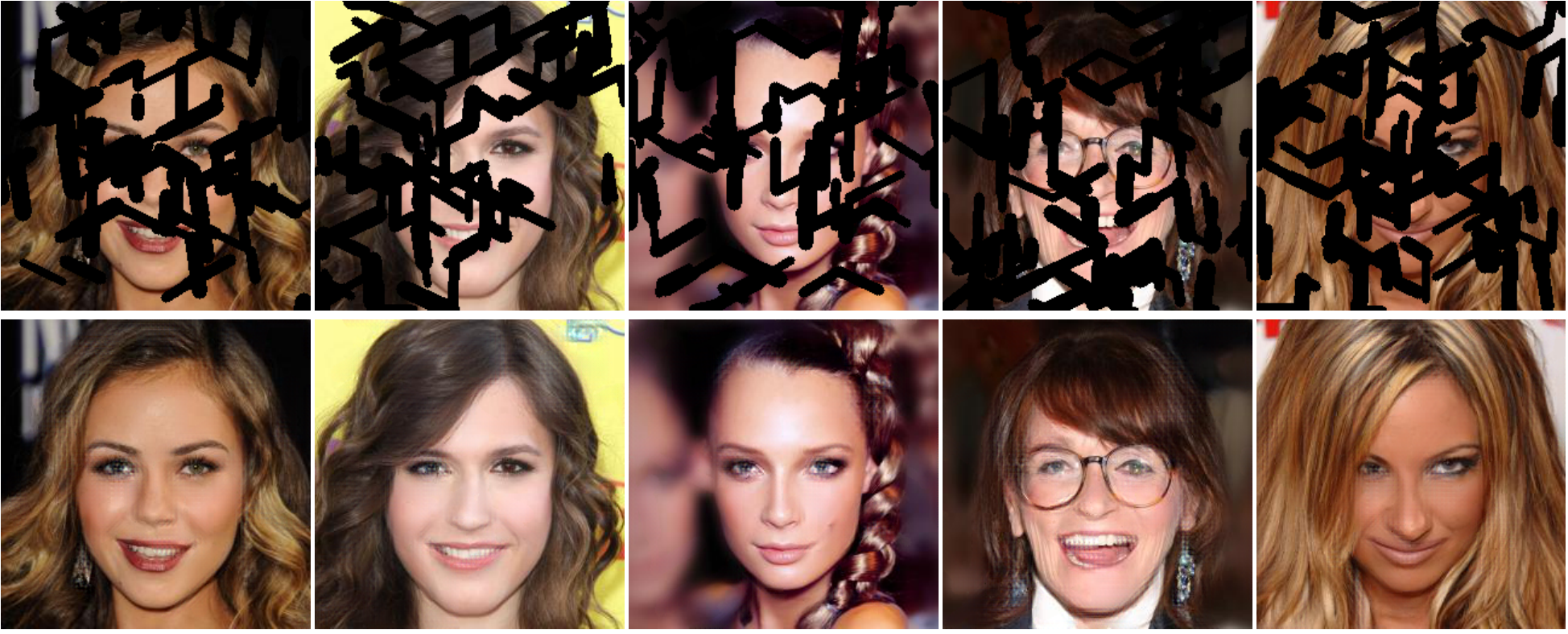}
\caption{Inpainted images (bottom row) generated by WavePaint from masked images using narrow masks (top row) from ImageNet validation set}
\label{fig:narrow}
\end{figure*}

We compare our models with the other state-of-the-art baselines as shown in Table~\ref{tab:celeb} for CelebA-HQ dataset. We compare the performance of the WavePaint across narrow, medium and wide masks. WavePaint consistently outperforms most of the other models, on a variety of mask configurations. It has to be noted most of the other models have much larger parameter count and employ adversarial training using a discriminator. Since WavePaint does not employ a discriminator it is light-wight, it can be trained faster than GANs and diffusion models. 

We could not compare WavePaint with latest diffusion models such as RePaint~\cite{lugmayr2022repaint} becasue diffusion is a much slower process of image generation and we were constrained in computational resources. RePaint~\cite{lugmayr2022repaint} had reported that quantitative results of LaMa~\cite{Lama_2021} are better than that of RePaint in wide and narrow mask inpainting on ImageNet and CelebA-HQ datasets. 

Since LaMa was a resource-efficient model for inpainting, we compared WavePaint with LaMa~\cite{Lama_2021} in  Table~\ref{tab:efficient} to analyse its resource-efficiency. We see that WavePaint requires less than one-fifth of the parameters of LaMa to outperform it in FID metric. WavaPaint is also $\sim 3 \times$ faster 
than LaMa in both inference and training speed and utilizes less than half the GPU consumed by LaMa. Our results clearly shows that WavePaint is more resource and parameter-efficient than LaMa. The high resource-efficiency of WavePaint can be attributed to the resource-efficient token-mixing using WaveMix blocks which processes the image at a lower resolution due to lossless downsampling property of 2D-DWT. The quantitative performance of WavePaint using different hyperparameters on CelebA-HQ and ImageNet datasets are shown in Table~\ref{tab:ablation}.

Table~\ref{tab:levels} shows the performance of WavePaint which uses WaveMix blocks with multi-level 2D-DWT. Using higher levels of DWT can improve the performance of the model due to the exponential increase in receptive field. 

\subsection{Qualitative Results}

The images generated by WavePaint on ImageNet dataset are shown in Figure~\ref{fig:picsimage}. We can see that WavePaint completes textures and missing details by completing the lines and filling in details. The images generated by WavePaint for wide, medium and narrow masks are shown in Figure~\ref{fig:wide}, Figure~\ref{fig:medium} and Figure~\ref{fig:narrow} respectively. WavePaint can fill in missing details of facial features, colour, texture, eyes and eyebrows even if major parts of the image are masks.

\section{Ablation Studies}

Multiple ablation experiments were conducted to optimize the network hyper-parameters and understand the utility of the network components. Table~\ref{tab:module} shows the performance of WavePaint with 8 WaveMix blocks arranged in different number of modules. Results shows that having less number of modules with large number of WaveMix blocks is more parameter-efficient but results in poor performance. When we decrease the number of WaveMix blocks in each module and increase the number of modules, the model become larger with higher parameter count. Modules with 4 Waveblocks each retain parameter-efficiency without degrading performance.

\begin{table}[]
\centering
\begin{tabular}{@{}rrrr@{}}
\toprule
\#Modules & \begin{tabular}[c]{@{}r@{}}\#WaveMix Blocks\end{tabular} & \#Params & LPIPS \\ \midrule
1 & 8 & 3.0 M & 0.085 \\
2 & 4 & 3.3 M & 0.079 \\
4 & 2 & 4.0 M & 0.079 \\ \bottomrule
\end{tabular}
\vspace{2mm}
\caption{Performance of WavePaint with 8 WaveMix blocks by varying the number of modules. Experiment was done on a subset of ImageNet dataset.}
\label{tab:module}
\end{table}

Removing DepthConv block from WavePaint reduces the FID score by 38\% and increases the training and inference throughput by 14\%. Since, depth-wise convolution is a highly parameter efficient operation, its removal only reduces the number of parameters by less than 1\%. Therefore, adding DepthConv block in each module is beneficial for the network as it aids the WaveMix block with further spatial token-mixing.

\section{Conclusion and Future Work}

This paper proposes using multi-level 2D-DWT token-mixing for the less explored task of image inpainting. The performance of the proposed model is comparable to much larger models and those that uses adversarial training on CelebA-HQ dataset. Also, our model uses only a fraction of the parameters, consumes less GPU RAM and is multiple times faster in training and inference compared to other models such as LaMa~\cite{Lama_2021}. A possible direction of future work is to develop resource-efficient image generation models using WavePaint trained in an adversarial or diffusion setting. Thus, this paper points to the the potential of using token-mixing as alternative to vision transformers and CNNs for resource-efficient image inpainting without the need for slower complex training procedures like adversarial and diffusion. The faster receptive field expansion leading to availability of global context information can help these models do image reconstruction on par with transformers.

\bibliographystyle{unsrt}  
\bibliography{references}

\end{document}